\documentclass[acmlarge]{acmart}

\AtBeginDocument{%
 \providecommand\BibTeX{{%
 \normalfont B\kern-0.5em{\scshape i\kern-0.25em b}\kern-0.8em\TeX}}}

\setcopyright{acmcopyright}
\copyrightyear{2023}
\acmYear{2023}
\acmDOI{XXXXXXX.XXXXXXX}

\newif\ifdraft
\drafttrue

\usepackage{amsmath}
\usepackage{url} 
\usepackage{rotating}
\usepackage{graphicx}
\usepackage{lscape}
\usepackage{subcaption}
\usepackage{makecell}
\usepackage{xcolor,pifont}
\usepackage{ragged2e}
\usepackage{balance}
\usepackage{caption}
\usepackage{booktabs}
\usepackage{listings}
\usepackage{multirow, multicol}

\newcommand{\killpunct}[1]{}

\begin{document}

\title{Explainable Authorship Identification in Cultural Heritage
Applications: Analysis of a New Perspective}

\author{Mattia Setzu} \email{mattia.setzu@unipi.it}
\orcid{0000-0001-8351-9999} \affiliation{%
\department{Dipartimento di Informatica} \institution{Università di
Pisa} \postcode{56127} \city{Pisa} \country{IT}}

\author{Silvia Corbara} \email{silvia.corbara@sns.it}
\orcid{0000-0002-5284-1771} \affiliation{ \institution{Scuola Normale
Superiore} \postcode{56126} \city{Pisa} \country{IT}}

\author{Anna Monreale} \email{anna.monreale@unipi.it}
\orcid{0000-0001-8541-0284} \affiliation{%
\department{Dipartimento di Informatica} \institution{Università di
Pisa} \postcode{56127} \city{Pisa} \country{IT}}

\author{Alejandro Moreo} \email{alejandro.moreo@isti.cnr.it}
\orcid{0000-0002-0377-1025} \affiliation{ \department{Istituto di
Scienza e Tecnologie dell'Informazione} \institution{Consiglio
Nazionale delle Ricerche} \postcode{56124} \city{Pisa} \country{IT}}

\author{Fabrizio Sebastiani} \email{fabrizio.sebastiani@isti.cnr.it}
\orcid{0000-0003-4221-6427} \affiliation{ \department{Istituto di
Scienza e Tecnologie dell'Informazione} \institution{Consiglio
Nazionale delle Ricerche} \postcode{56124} \city{Pisa} \country{IT}}

\renewcommand{\shortauthors}{Setzu et al.}

\begin{abstract}
  \noindent While a substantial amount of work has recently been
  devoted to enhance the performance of computational Authorship
  Identification (AId) systems, little to no attention has been paid
  to endowing AId systems with the ability to explain the reasons
  behind their predictions. This lacking substantially hinders the
  practical employment of AId methodologies, since the predictions
  returned by such systems are hardly useful unless they are supported
  with suitable explanations. In this paper, we explore the
  applicability of existing general-purpose eXplainable Artificial
  Intelligence (XAI) techniques to AId, with a special focus on
  explanations addressed to scholars working in cultural heritage. In
  particular, we assess the relative merits of three different types
  of XAI techniques (feature ranking, probing, factuals and
  counterfactual selection) on three different AId tasks (authorship
  attribution, authorship verification, same-authorship verification)
  by running experiments on real AId data. Our analysis shows that,
  while these techniques make important first steps towards
  explainable Authorship Identification, more work remains to be done
  in order to provide tools that can be profitably integrated in the
  workflows of scholars.
\end{abstract}


\maketitle


\section{Introduction}
\label{sec:intro}

\noindent \textit{Authorship Analysis} can be broadly defined as ``any
attempt to infer the characteristics of the creator of a piece of
linguistic data''~\citep[p.\ 238]{Juola:2006fk}, where these
characteristics include the author’s biographical information (e.g.,
age group, gender, mother tongue, etc.) and identity. Since the
pioneering work of \citet{Mosteller:1963fr}, the field of authorship
analysis has made extensive use of computational methods, thereby
contributing to the work of many scholars in the field of cultural
heritage and providing them with new tools and perspectives in the
study of historical documents of different languages and periods.

One important group of tasks in authorship analysis goes under the
name of \textit{Authorship Identification} (AId), and concerns the
study of the true identity of the author of a written document of
unknown or disputed paternity. The three main tasks in the AId group
are \textit{Authorship Attribution} (AA), \textit{Authorship
Verification} (AV), and \textit{Same-Authorship Verification}
(SAV). In AA~\citep{Koppel:2009ix, Stamatatos:2009ye}, given a
document $d$ and a set of candidate authors
$\{A_{1}, \ldots, A_{m}\}$, the goal is to identify the most likely
author of $d$ among the set of candidates; AA is thus a single-label
multiclass classification problem, where the classes are the authors
in $\{A_{1}, \ldots, A_{m}\}$.\footnote{In classification,
\textit{multiclass} (as opposed to \textit{binary}) means that there
is a set of $m>2$ classes to choose from; there are instead just 2
classes to choose from in the binary case.
} In AV~\citep{Stamatatos:2016ij}, given a candidate author $A$ and a
document $d$, the goal is to infer whether $A$ is the real author of
$d$ or not; AV is thus a binary classification problem, with $A$ and
$\overline{A}$ as the possible classes. In SAV~\citep{Corbara:2023kk},
given two documents $d_{1}$ and $d_{2}$, the goal is to infer whether
they are written by the same (possibly unknown) author or not; SAV is
thus also a binary classification problem, with \textsc{SameAuthor}
and \textsc{DifferentAuthor} as the possible classes.
All of these tasks are usually approached as \emph{text
classification} tasks, whereby a supervised machine learning
algorithm, using a set of labelled documents, is used to train a
classifier to perform the required prediction task.

A close analysis of the AId literature reveals that, while researchers
have devoted a lot of effort to test the relative performance of
different learning methods in AId tasks, to check the usefulness of
different types of features for capturing written style, and to apply
the techniques thus developed to a number of AId case studies, little
to no attention has been paid to providing users with
\emph{explanations} regarding the predictions of the above
algorithms. This is unsatisfactory, since machine-learned classifiers
are usually opaque (i.e., they provide predictions but do not provide
intuitive explanations of the reasons behind these predictions), and
most users of AId systems
hardly assign any value to a ``bare'' automated prediction, and are
instead interested in knowing the \textit{reason} behind the system
prediction.

The goal of this work is to make progress towards filling this gap, by
carrying out an in-depth analysis of the suitability to the three main
AId tasks of a set of well-known general-purpose \emph{eXplainable
Artificial Intelligence} (XAI) methods, i.e., methods for explaining
the predictions of a machine-learned system. In doing this, the users
of AId systems that we have in mind are scholars working in cultural
heritage (such as philologists, historians, linguists), who are
typically not machine learning experts. Note that, in this research,
our goal is not to devise a new XAI method, but to examine the
suitability of existing XAI methods to AId tasks and to the user group
identified above.

This paper is organised as follows. After a discussion on
(computational) AId and on the importance of explanations for the
predictions issued by machine-learned AId systems
(Section~\ref{sec:background}), in Section~\ref{sec:related} we survey
relevant related work. In Section~\ref{sec:XAID} we explain the three
major classes of methods for explaining the predictions of
machine-learned systems that we explore in this paper, i.e., feature
ranking, transformer probing, and factuals and counterfactuals
selection. In Section~\ref{sec:setup} we explain our experimental
setup, while in Section~\ref{sec:results} we showcase the application
of the aforementioned methods to AId tasks, and analyse their relative
benefits for the specific purposes within the cultural heritage
domain. Section~\ref{sec:conclusion} concludes, pointing at avenues
for future research.


\section{Background: Authorship identification and the need for
explanations}
\label{sec:background}

\noindent As mentioned in the introduction, AId tasks are usually
tackled as text classification problems~\citep{Aggarwal:2012wl}, and
solved by using supervised machine learning algorithms. For instance,
in order to solve the AV task, a machine learning algorithm trains a
binary ``$A$ vs.\ $\overline{A}$'' classifier using a training set of
labelled texts, where the training examples labelled as $A$ are texts
by the candidate author and the training examples labelled as
$\overline{A}$ are texts by other (ideally, stylistically similar)
authors.
 
Generally speaking, AId techniques attempt to spot the ``hand'' of a
given writer, thus distinguishing their written production from the
production of others. The core of this practice, also known as
``stylometry''~\citep{Holmes:1998en}, does not rely on the
investigation of the artistic value or the meaning of the written
text, but on a quantifiable characterisation of its style. This
characterisation is typically achieved via an analysis of the
frequencies of linguistic events (also known as ``style markers'')
occurring in the document of interest, which are assumed to remain
more or less constant throughout the production of a given author,
while conversely varying substantially across different
authors~\citep[p.\ 241]{Juola:2006fk}. These linguistic events are
often of seemingly minimal significance (such as the use of a
punctuation symbol or a preposition), but are assumed to be out of the
conscious control of the writer, and hence to occur in patterns that
are hard to consciously modify or imitate.

AId methodologies are profitably employed in many fields, ranging from
cybersecurity~\citep{Schmid:2015qq} to computational
forensics~\citep{Chaski:2005pd, Larner:2014kl, Perkins:2015rp,
Rocha:2017yy}; yet another important area of application for AId
techniques is the cultural heritage field, which is the focus of the
present article. Indeed, researchers might use AId techniques to infer
the identity of the authors of texts of literary or historical value,
whose paternity is unknown or disputed. In these cases, unknown or
disputed authorship may derive from authors attempting to conceal
their identity (whether for a desire to remain anonymous or for the
malicious intent to disguise themselves as someone else), or simply as
a result of the passing of time, which is a common occurrence when
dealing with ancient texts~\citep{Kabala:2020bu, Kestemont:2016fh,
Savoy:2019qr, Stover:2016zl, Tuccinardi:2017yg}.
 
While many efforts in AId have focused on testing the accuracy of
different learning algorithms (see for example the surveys by
\citet{Stamatatos:2009ye, Juola:2006fk, Grieve:2007ti}, or the annual
editions of the popular PAN shared task~\citep{Stamatatos:2022or,
Kestemont:2021io}), or on proposing new sets of features that these
algorithms could exploit \citep{Corbara:2023kk, Sari:2018fe,
Wu:2021fd}, or simply on applying known techniques to case studies of
literary interest \citep{Kabala:2020bu, Kestemont:2016fh,
Savoy:2019qr, Stover:2016zl, Tuccinardi:2017yg}, little or no effort
has been devoted to endowing these systems with the ability to
generate explanations for their predictions.

This fact represents indeed a very important gap in the literature,
and a hindrance to a more widespread adoption of these technologies in
cultural heritage and other fields. The ability to provide
justifications for their own predictions is a very important property
for machine-learned systems in general, and even more so when these
systems are involved in significant decisions-making processes, such
as deciding on the authorship of written documents, with all its legal
and ethical implications. We might even claim that \emph{an authorship
analysis system is almost useless, unless it is endowed with the
ability to explain its own decisions}. Indeed, when such a system is
applied to, say, determining the authorship of an important literary
work of controversial paternity~\citep{Corbara:2020qc, Corbara:2022kt,
Kestemont:2015lp, Tuccinardi:2017yg}, it is paramount that the
prediction is presented to the domain experts along with a
comprehensive explanation of the reasons why the system made such a
prediction. There are two main reasons for this.
 
The first one is that a domain expert who has devoted a sizeable
intellectual effort to determining the authorship of a given document
is unlikely to blindly trust the prediction of an automatic system,
unless the possibility to examine the reasons of its prediction and/or
the inner working of the system is
provided~\citep{Ribeiro:2016jn}. Indeed, a domain expert might want to
check whether the AId system is actually focusing on the writing style
of the document under investigation (and on features deemed important
by the expert), and that the system is not instead focusing on other
possibly misleading aspects of the document, such as its topic. A
similar argument can be applied to an automated prediction meant to be
used as evidence in a criminal case: in this case, it would be
necessary to put the judge and the jurors in the condition to form
their own opinion regarding the output of the automatic system, by
giving them as much information as possible on the system and on the
reasons that have led it to make that specific
prediction~\citep{Bromby:2011ji, Larner:2014kl, Halvani:2021nw}.

The second reason is that, in the case of cultural heritage
applications, the knowledge regarding the process of an AId system
might inspire the domain expert with new possible working hypotheses
that had not been considered before (e.g., by highlighting a
linguistic event that prominently occurs in one author's works but not
in the production of other authors). In this regard, it is interesting
to note that, in authorship analysis studies, the domain expert and
the automatic system often employ complementary methodologies. For
instance, when performing authorship analysis for cultural heritage
texts, a domain expert may (i) analyse the historical facts described
in the text and check whether a certain candidate author could
possibly have been aware of these facts; (ii) analyse the stand that a
candidate author takes towards a certain issue, and check whether this
stand is compatible with what we already know about the author's
ideas; and (iii) in general, bring to bear their knowledge of a given
candidate author, of the historical period in which the candidate
operated, of the cultural \textit{milieu} that surrounded the
candidate, and decide whether all these are, or are not, compatible
with the hypothesis that the candidate may be the real author of the
disputed document. Current automatic AId systems can do none of the
above. More in general, while the domain expert can use
\emph{exogenous} real-world knowledge (i.e., knowledge external to the
document), an automatic AId system is typically only able to use
\emph{endogenous} knowledge (i.e., knowledge extracted from the
document -- plus potentially some external linguistic knowledge, in
the form of dictionaries, or sets of word embeddings, or
similar). However, an automatic system is capable of doing
fine-grained statistical analyses that would be difficult, or
impossible, for any human to perform;\footnote{Domain experts
sometimes do analyse the same features as automatic systems, e.g.,
they may notice that an author tends to use a specific spelling of a
given word, or that an author tends to start a sentence with a certain
word or sequence of words. However, it is undeniable that a human
carries out this type of analyses with greater difficulty and only on
a limited scale.} stylometric analysis is indeed one such type of
analysis, where an automatic system can analyse a huge amount of
linguistic traits of apparently minimal significance that, altogether,
can define an author's style. In other words, this ``lower-level''
analysis of the text provides a useful complement to the
``higher-level'' analysis that the domain expert carries out.

To summarise, the role of an automatic system in tasks such as AId
should not be that of an opaque, cryptic oracle, but that of a tool
that supports the domain expert, who is in charge of delivering the
final authorship hypothesis. In other words, the automatic system
should be integrated within a pre-existing workflow; by doing so, it
could be perceived not as an attempt to replace the domain experts,
which would understandably elicit a negative reaction on their part,
but as an attempt to support them in their job.
 
There are three main obstacles in devising an explainable AId
system. First, the vector space typical of text-related prediction
tasks usually has a very high dimensionality; indeed, many of the
tools that have been developed in the XAI literature are more suited
to the low dimensionality typical of structured data. Second, the
linguistic events employed as features in AId tasks are usually of
minimal significance (e.g., the occurrence of a specific character
3-gram), a significance that may be hard to grasp for the person to
whom the explanation is addressed; this is indeed an intrinsic problem
stemming from the different approaches that humans and machines employ
when facing AId tasks. The third obstacle (which is inherently related
to the first two) is that, in text-related prediction tasks, a
prediction is obtained thanks to the contribution of \emph{many}
features, all representing linguistic events of minor importance; in
other words, it is difficult to isolate one or few such events that
are responsible for the final prediction by themselves. Moreover, as
noted by \citet{Halvani:2021nw}, the bag-of-features representation,
which is usually employed in AId tasks, loses the contextual
information of the individual features, making it difficult to
understand how such features relate to each other with regard to the
final output. This means that presenting the user with a concise
explanation of the prediction (in terms of the features that have
contributed to it) is usually a very difficult matter.


\section{Related work}
\label{sec:related}

\noindent In recent years, XAI has gained more and more attention in
the NLP and text mining communities; see for example the general
surveys on XAI by \citet{Hamon:2020ss, Guidotti:2018ye,
Carvalho:2019hi, Linardatos:2020pl}, the surveys on XAI applied to NLP
and text classification by \citet{Lertvittayakumjorn:2019hu,
Danilevsky:2020sy}, and the recent proposals discussed in the works of
\citet{Rajagopal:2021va, Wiegreffe:2019tt, Liu:2019sw, gu2021dream}.

XAI methods are usually divided into \emph{local explainers} and
\emph{global explainers}. A local explainer is a method that returns
an explanation for a specific prediction of the classifier, while a
global explainer explains the behaviour of a classifier in general,
with no reference to a specific prediction. Understandably, each
approach has its own pros and cons, but both can be used to offer
insight in the rationale of a classification decision. Since the two
approaches focus on different kinds of information, they can be
complementary, and multiple local explanations can be combined to gain
a general understanding of the behaviour of the classifier
\citep{Boenninghoff:2019be, lundberg2020local}.

\begin{figure}
  \includegraphics[width=\textwidth]{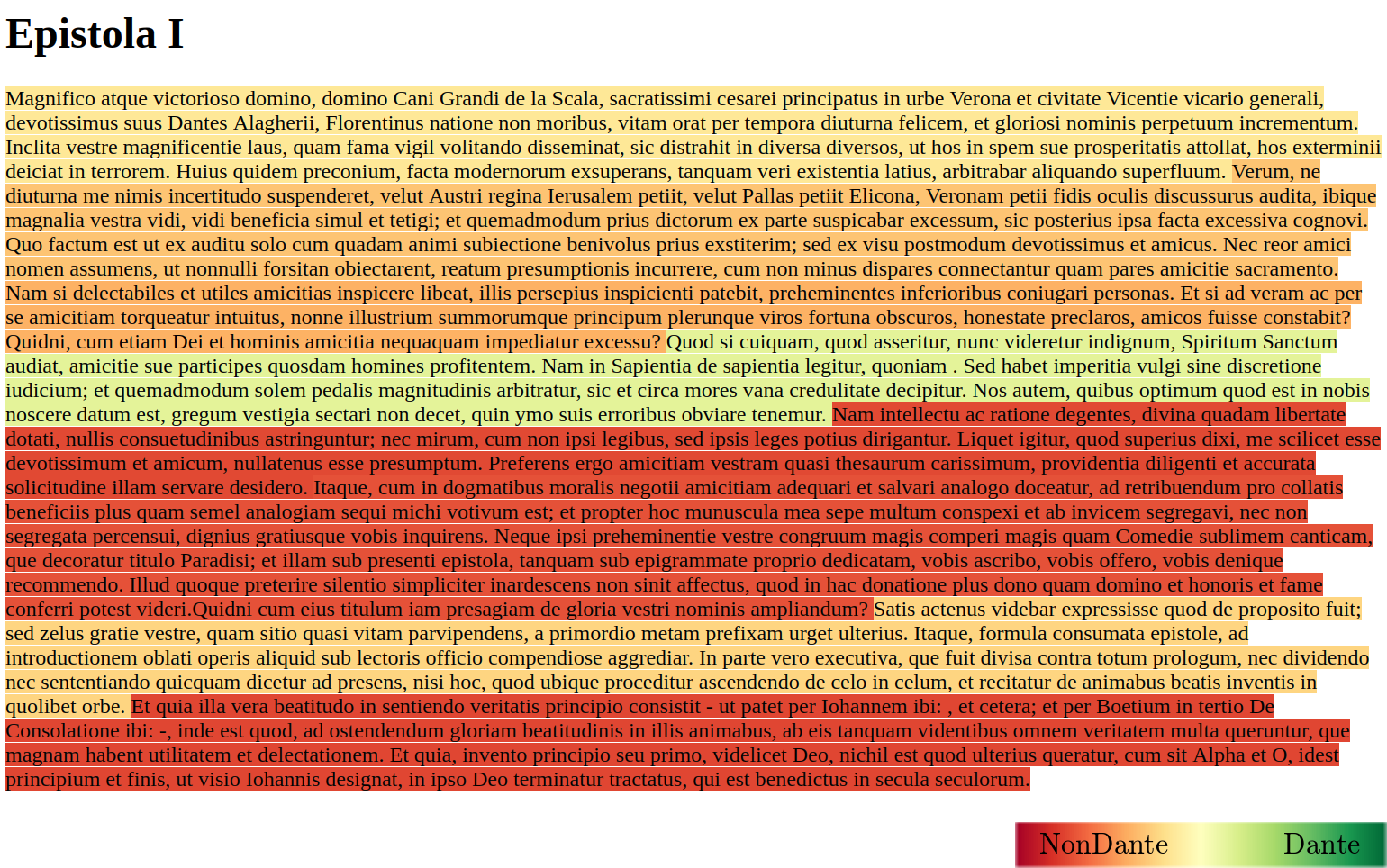}
  \caption{\label{fig:epistle} Visualisation of the \textit{Epistle to
  Cangrande} (from \citet{Corbara:2020qc}), whose attribution to Dante
  Alighieri is uncertain; paragraphs on the red side of the spectrum
  are those that the authorship classifier believes to be ``less
  Dantean'', while those on the green side of the spectrum are those
  that the authorship classifier believes to be ``more Dantean''.}
\end{figure}

Despite the growing interest that XAI has witnessed in recent years,
little or no attention has been given to its application to AId,
possibly also due to the difficulties mentioned at the end of
Section~\ref{sec:background}. Some recent attempts towards providing
explanations for the predictions of text classifiers consist of
creating a saliency mask \citep{Guidotti:2018ye}, as in, visually
displaying the textual elements that have proven most important for
the decision of the classifier directly within the document (this is
thus an example of a \emph{local} explainer). For example, when
tackling AV for the \textit{Epistle to Cangrande}, a medieval Latin
text traditionally attributed to Dante Alighieri,
\citet{Corbara:2020qc} highlight the 90 paragraphs of the
\textit{Epistle} with different colours based on the classification
scores obtained when classifying each paragraph individually (see
Figure~\ref{fig:epistle}); here, the score obtained by a paragraph is
thus used as an approximation of how much a paragraph contributes to
the final prediction for the entire document.
\citet{Theophilo:2022la} obtain a similar effect at the feature level
by adapting the popular LIME algorithm\footnote{Specifically, given a
complex model $\Phi$ and an instance $x$, LIME \citep{Ribeiro:2016jn}
employs a perturbation algorithm that generates a neighbourhood of
$x$. Leveraging this neighbourhood and the prediction made by $\Phi$
on said neighbourhood, LIME learns a new linear classifier that is a
good approximation of $\Phi$ (i.e., it outputs similar
predictions). This linear classifier is intrinsically interpretable,
providing coefficients for each input feature, hence allowing the user
to understand what features have contributed most to the prediction of
$\Phi$ on $x$. Note that the original LIME formulation for text is
restricted to word and character unigrams as interpretable
components.} to process character 4-grams. In order to offer an
explanation for the decisions of his compression-based SAV algorithms,
\citet{Halvani:2021nw} proposes to colour the two texts based on
their differences (the higher the discrepancy, the stronger the
colour), thus providing an intuitive and straightforward
representation of areas of the texts that play a more important role
in the prediction.\footnote{\citet{Halvani:2021nw} also proposes to
display the element-wise Manhattan distance between the two values of
the same feature, which represents
how much the feature influences the similarity of the two documents.}
Alternatively, when working with architectures based on neural
networks, researchers have focused on the visualisation of the
attention weights~\citep{Boenninghoff:2019be}, or by computing the
derivative of the output given the embedding of a word in the input
\cite{Shrestha:2017cn}.

While saliency maps and similar visualisation devices may help the
user to focus on areas of the text that have played an important role
in the system decision, they are incomplete explanations, since they
place on the user the burden of understanding \emph{why} the system
has reached exactly that decision. An alternative method consists of
ranking the features used by the classifier by their importance (this
is thus an example of a \emph{global} explainer), where this
``importance'' can be assessed in different ways. For example, in
their work on native language identification (the task of detecting
the native language of the author of a text), \citet{Berti:2023mw} use
the weights associated to the features in a linear classifier as
indications of which features best separate the classes, since the
absolute value of these weights is proportional to the discriminative
power of the respective features.\footnote{E.g., in the
\textsc{Spanish} vs.\ \textsc{NonSpanish} classifier, the weight of
\textit{especial}, a misspelling of the English word \textit{special},
is high and positive, leading to the class \textsc{Spanish}, since
native speakers of Spanish have a tendency to prefix a spurious
\textit{e-} to many English words starting with an \textit{s}, due to
an interference from their mother tongue. As a result, when a text
classified as \textsc{Spanish} contains the term \textit{especial},
this occurrence constitutes a (partial) explanation of this
classification decision.} Other studies, such as the one by
\citet{Sapkota:2015nt}, assess the effect of different feature types
(e.g., character $n$-grams) by evaluating the performance of a
classifier trained without the feature types under study. For neural
networks, and particularly for CNNs, an approach similar to the above
consists of listing the input elements that generate the highest
activation values aggregated over all filters, or the input elements
that generate a significant activation value for the highest number of
filters~\citep{Shrestha:2017cn}. However, these approaches are
admittedly a long way from constituting satisfactory explanations for
AId decisions, because they provide explanations that are partial
and/or difficult to grasp for a scholar who is not a machine learning
expert.

\begin{figure}[t!]
  \begin{center}
    \includegraphics[width=\textwidth]{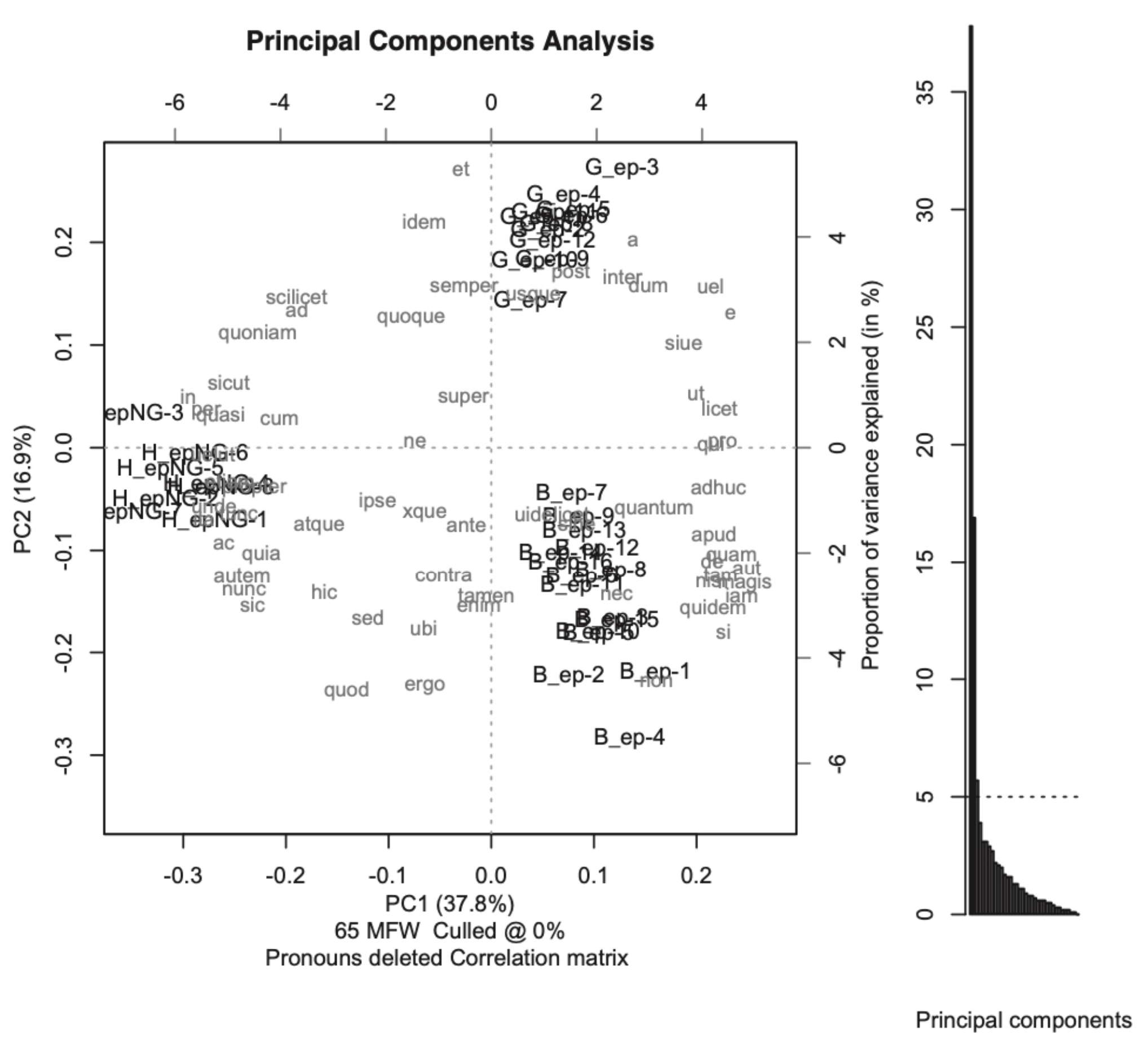}
    \caption{\label{fig:KestemontPCA}Visualisation (from
    \citet{Kestemont:2015lp}), obtained via PCA, of texts from three
    different authors (here identified by the letters B, G, H),
    showing that the technique used separates them well; strings
    $A$\_ep-$n$ identify the $n$-th text by author $A$ in the
    dataset. The words that are located near the texts by author $A$
    are the ones that occur more frequently in the texts by $A$ than
    in the texts by the other two authors.}
  \end{center}
\end{figure}

Another widely used technique consists of displaying the documents of
interest in a bidimensional space obtained through dimensionality
reduction (e.g., via principal component analysis), in order to
provide a visual idea of the characteristics of the data
\citep{Binongo:2003wt, Forstall:2011pu, Kestemont:2015lp}. As an
example, in Figure~\ref{fig:KestemontPCA} texts are mapped to a
bidimensional space together with the words whose use most
differentiates the candidate authors. A reader is thus able to see how
texts by the same author are clustered together, and how the
classifier has found the use of specific words to be idiosyncratic to
specific authors.

\section{Methodology}
\label{sec:XAID}

\noindent As briefly discussed in Section~\ref{sec:related}, there are
several general-purpose XAI methodologies that allow to better
understand a trained classifier and/or a specific prediction. In this
work, we experiment with three of these options, analysing their
suitability to AId tasks in general, and to the specific public of
cultural heritage professionals in particular. Within this context, we
discuss the possible contribution of both global explainers and local
explainers.

In Section~\ref{subsec:XAID_feats_ranking} we show how to gain insight
into the features that a linear classifier deems most important for
the classification task; we do so by directly employing the weights of
the trained model, and show how to obtain both global and local
explanations. Note that, in the case of linear classifiers, and more
generally in the case of ``classic'' machine learning methods (such as
SVMs, logistic regression, decision trees, shallow neural networks),
the features used to train the learning algorithm are identified
\textit{a priori} by the researcher, in the so-called ``feature
engineering'' phase. In other words, the model, and thus the
explanation, is constrained to use only the features (and combinations
thereof) defined by the researcher.

Conversely, deep neural networks are by design able to discover novel
discriminative features from the data, and can thus carry out the
feature engineering phase autonomously.  While some feature ranking
solutions for deep-learning models are present in the literature
\citep{Lundberg:2017op}, they are extremely sensitive to the input
data, and are based on a number of assumptions (such as uniform data
distribution) which are often unrealistic \citep{Hooker:2019zl}. Thus,
in the case of AId solutions based on deep learning, rather than
providing hardly interpretable and/or unreliable explanations, we will
thus employ a more sophisticated solution, known as ``model probing'',
as applied to a RoBERTa-based model, in order to obtain global
explanations (see Section~\ref{subsec:XAID_probing}).

Finally, in Section~\ref{subsec:XAID_factuals} we discuss how to
extract prototypical examples from the training set. These
representative examples provide the user with instances that the model
deems most similar to the test example and that belong to the same (or
different) class as the model prediction, exposing the internal
representation learned by the model, and thus acting as local
explainers.


\subsection{Feature ranking}
\label{subsec:XAID_feats_ranking}

\noindent A common strategy for offering a global explanation (i.e.,
providing a general understanding of the behaviour of the classifier)
is to show the features the classifier mostly focuses on during
prediction, as already explained in Section~\ref{sec:related}. In
these XAI methods, given a trained model, each feature is associated
to a score, and the employed features are presented in decreasing
order based on their score.

The score of a feature can be obtained in various ways. In the case of
linear models, the most direct way is to employ the coefficients (or
weights) of the classifier. By design, a linear classifier has the
form $h: \mathbf{x}\cdot\mathbf{w} + b$, where $\mathbf{x}$ is the
feature vector that represents data item $x$, $\mathbf{w}$ is a vector
of weights learned from the training data (one weight for each
feature), and $b$ is the intercept of the function; item $x$ is
assigned the positive class when $h(\mathbf{x})>0$, and it is assigned
the negative class otherwise. In the application scenario we discuss
in Section~\ref{subsec:setup_methods}, where the feature vectors fed
to the linear SVM are positive definite, the higher the absolute value
of the weight $w_{i}$ associated to the $i$-th feature, the larger is
the contribution of such feature towards the prediction.

Note that linear methods compute a set of coefficients for each binary
classification problem (which is the case of SAV and AV). In the case
of multiclass classification (which is the case of AA), the learner
computes a set of coefficients for each class; for prediction
explanation purposes, these sets must be examined individually.

The coefficients of the model can also be used to obtain a form of
local explanation: by multiplying the feature value extracted from a
test example by the correspondent coefficient, we can assess how much
the feature determined the specific prediction for the document. This
allows us to understand the model both on a global level and on a
local level, explaining both how the model \emph{generally} reasons,
and how it reasons on \emph{specific instances}.

There are more sophisticated ways to get feature scores (and they are
a mandatory resource in the case of non-linear methods, such as neural
networks). For example, SHAP~\citep{Lundberg:2017op} is a widely used
family of algorithms for model-agnostic XAI (meaning that it can be
applied to any learning algorithm). Unlike LIME, SHAP performs
perturbations on the feature set, then queries the model to estimate
the importance of each feature by leveraging the change in prediction
that each perturbation has produced on the outcome of the model. By
default, SHAP scores are local explanations, but the scores from
multiple examples can be averaged to reach a global
explanation. However, since the number of features employed in textual
settings is usually extremely high, the number of perturbations that
the SHAP algorithm should compute would be exponentially large, making
it computationally prohibitive; in these cases, perturbations can be
approximated through random sampling, but this is only a band-aid
solution.

Even though in this case the features employed by the classifier are
defined \textit{a priori}, an explanation of the type described above
can be extremely useful for the scholar. For instance, in the tens of
thousands character $n$-grams that can be extracted from the texts,
what are the most discriminative for the author(s) of interest? Thanks
to the explanations mentioned above, in theory a scholar might find
out, for example, that a certain author tends to avoid certain
patterns of characters, or vice versa has a preference for specific
syntactic constructs.


\subsection{Probing}
\label{subsec:XAID_probing}
 
\noindent As already shown, obtaining an indication of the importance
of the features by using the feature weights of the model is
straightforward for linear classifiers; however, it is not as
straightforward for non-linear classifiers, such as the ones
exploiting neural networks. Nevertheless, explainability is even more
important for these ``black-box'' architectures, for at least two
reasons. On the one hand, since the features are not identified
\textit{a priori} by the designer (as it is instead the case with
``traditional'' learners), an explanation method may allow the scholar
to check if the classifier is using the features that they indeed deem
important for the recognition of authorial style, and thus it might
help them to trust the classification system (see
Section~\ref{sec:background}). On the other hand, an explanation
method may allow the scholar to check if the system has discovered new
features that are interesting for identifying the authorship of
written documents, and that can be interesting to investigate further.

Indeed, many recent studies have tackled the far-from-trivial task of
developing XAI techniques that can show what features these models are
actually leveraging in their predictions. Among these studies, the
method of ``probing'' has recently gained vast popularity
\cite{Belinkov:2022hn}. Probing allows a user to understand if a
certain feature of interest (not defined \textit{a priori} by the
designer) has been learned and used by the model. For instance,
probing has been used to discover that some famous pre-trained
language models, such as BERT and RoBERTa, are not really capable of
understanding basic mathematical concepts \citep{Lin:2020ii}, but seem
to have learnt some form of common sense directly from data
\citep{Jullien:2022st}. The main idea behind the process of probing is
to input the latent representation computed by the neural network
model (from now on, the \emph{main model}) to a second, very simple
model (from now on, the \emph{probe}), whose task is to predict
whether the feature of interest is present in the latent
representation or not. Given the simplicity of the probe and the
complexity of the representation, the underlying assumption is that,
if the presence of a feature can be found even with a simple probe,
then that feature is encoded by the main model in the latent
representation.

Specifically, given a non-linear model $\Phi$ and an hypothesis
feature $f$, in order to probe the model (that is, when trying to
provide an answer to the question ``Does $\Phi$ internally learns from
$f$?''), we create a dataset of the form
$\{\phi(x_i), f(x_i)\}_{i=1}^n$, in which $x_i$ is a textual document,
$\phi(x_i)$ is the internal representation of $x_i$ created by $\Phi$,
and $f(x_i)$ is a function that characterises $x_i$ in terms of the
feature $f$. For example, $f(x_i)$ may be binary, returning 1 or 0 to
indicate that a given feature is present or absent in $x_i$,
respectively. Conversely, $f(x_i)$ may be categorical, returning a
class label in the range $\{1, \ldots, n\}$ when the characteristics
of $f$ in $x_i$ allow us to distinguish amongst $n$ different groups
of documents (see Section~\ref{subsec:results_probing}).
We then train a linear model with this dataset, and we use the
resulting classifier to estimate (e.g., via cross-validation) the
extent to which the characteristics encoded by the feature under study
are directly learnable from the internal representation of $\Phi$. We
repeat this process for every feature we conjecture could be playing a
role in the decision function that the model implements.

In particular, in Section~\ref{subsec:results_probing}, we exemplify
this approach by developing five types of probing:

\begin{itemize}

\item \textbf{POS chains}: we probe the model for features extracted
  from the concatenation of Part-Of-Speech (POS) tags, which are
  nowadays a standard feature type for AId (see for example
  \citet{Jafariakinabad:2020sc});
\item \textbf{SQ chains}: we probe the model for features extracted
  from the concatenation of Syllabic Quantities (SQ), which have been
  first proposed for AId tasks in the Latin language by
  \citet{Corbara:2023kk}; \footnote{In the Latin language, words can
  be divided into syllables, which can be long or short depending on
  their quantity; see \citet{Corbara:2023kk} for more information.}
\item \textbf{Word lengths}: we probe the model for the frequency of
  word lengths, which have been employed as features in the AId field
  since the proposal by \citet{Mendenhall:1887cc};
\item \textbf{Function words}: we probe the model for the frequency of
  function words, which are widely employed as features in the AId
  field~\citep{Binongo:2003wt, Kestemont:2014uc};
\item \textbf{Doc genre}: we probe the model for the genre of the
  document, in order to see whether the model encodes the
  characteristics of the genre into the latent representation.
\end{itemize}

\noindent Given the generality of the probing approach, any feature
type can be easily that the domain expert may find interesting to
investigate can be probed.


\subsection{Selection of factuals and counterfactuals}
\label{subsec:XAID_factuals}

\noindent Given a prediction $\hat{y}$ on an item $\mathbf{x}$, it
might be useful for a domain expert to check what the classifier
considers similar items, in order for them to i) understand whether
the similarities estimated by the classifier indeed make sense
according to what the expert already knows about them
(e.g., the classifier considers documents from the same historical
period similar), and ii) discover possible similarities among the
written documents that the expert might have been unaware of, but the
classifier has brought to light.

To this aim, a standard method is to retrieve the training instances
that are most similar to $\mathbf{x}$ according to the model.  Among
these training items, some would have the same class as $\hat{y}$,
while others would have a class different from $\hat{y}$; the former
are called \textit{factuals}, and the scholar might find them useful
when trying to understand the characteristics of the predicted class,
while the latter are called \textit{counterfactuals}, and they might
be useful in allowing the scholar to gauge the minimal requirement for
the classifier to predict a class $y \neq \hat{y}$. The similarity of
two instances can be computed by applying any standard similarity
measure directly to the input vectors in the case of linear models
with no internal representation, or to the latent representations of
the instances in the case of deep learning models. In the case of a
linear model, it is also possible to easily spot the features that
most contributed to the similarity of the two items.


\section{Experimental setup}
\label{sec:setup}

\noindent In this work, we tackle the three major AId tasks, i.e.,
Authorship Attribution (AA), Authorship Verification (AV), and
Same-Authorship Verification (SAV). We show how some well-known XAI
methodologies deal with predictions issued in each of these three
tasks on a dataset of medieval Latin \citep{Corbara:2022kt}.

In the following paragraphs, we present our experimental setup, which
exemplifies common AId settings. In particular, in
Section~\ref{subsec:setup_dataset} we present the dataset we use,
while in Section~\ref{subsec:setup_methods} we explain our
classification methodology, along with the learning algorithms we
employ.

The Python code to replicate our experiments is available at:
\url{https://github.com/silvia-cor/XAId}


\subsection{Dataset}
\label{subsec:setup_dataset}

\noindent In this study, we employ the dataset developed by
\citet{Corbara:2022kt}. The authors originally divided it into two
sub-datasets, \textsf{MedLatinEpi} and \textsf{MedLatinLit}, both
containing works in medieval Latin prose, mostly dating to the 13th
and 14th centuries; \textsf{MedLatinEpi} is composed of 294 texts of
epistolary genre, while \textsf{MedLatinLit} is composed of 30 texts
of various nature, especially literary works and chronicles. For this
project we select only 5 authors: Dante Alighieri and Giovanni
Boccaccio (who have documents in both the sub-datasets), Pier della
Vigna (who is the most prolific author of \textsf{MedLatinEpi}),
Benvenuto da Imola and Pietro Alighieri (who are the two most prolific
authors remaining in \textsf{MedLatinLit}). We delete the quotations
from other authors and the parts in languages other than Latin, both
marked in the texts. Following \citet{Corbara:2023kk}, we also divide
each text into sentences, where a sentence is made of at least 5
distinct words (we attach shorter sentences to the next sentence in
the sequence, or to the previous one in case the sentence is the last
one in the document); we use each non-overlapping sequence of 10
consecutive sentences as a textual example. By doing this, we end up
with 2,729 text example in total. We randomly split the corpus into a
training set ($90\%$ of the examples) and a test set (remaining
$10\%$) in a stratified fashion.

For SAV, we do not employ all the pairs of segments that can be
created within the training and test sets, since their number would
quickly explode and would thus drastically slow down the
computation. In particular, given a set of authors
$\{A_{1}, \ldots, A_{z}\}$, we create $n$ \textsc{SameAuthor} pairs
for each author $A_{i}$ (each consisting of two random texts by
$A_{i}$), and $m$ \textsc{DifferentAuthor} pairs in total (where a
\textsc{DifferentAuthor} pair consists of two random texts by two
different authors in $\{A_{1}, \ldots, A_{z}\}$); the pairs are
unique. In our experiments we set $n$=5,000 and $m$=25,000 for both
the training set and the test set; therefore, both the training set
and the test set are balanced.

For the AV task, we select the author Dante Alighieri as the author of
interest, in line with the experiments in \citet{Corbara:2022kt}.


\subsection{Learning methods}
\label{subsec:setup_methods}

\noindent In this study, we experiment with offering explanations for
the output of one representative ``classic'' machine learning method
and for one representative deep learning method in the AId setting.

For the former, we employ a linear Support Vector Machine (SVM), a
very popular learner in AId tasks~\citep{Zheng:2006wf,
Kestemont:2019ov}; we use the implementation available from the
\textsf{scikit-learn} library~\citep{scikit-learn}. We fine-tune the
hyperparameter $C$ (with values in the range
$[0.001, 0.01, 0.1, 1, 10, 100, 1000]$) by performing 3-fold
cross-validation on the training set. In order to train the algorithm,
we compute the TfIdf values of all character n-grams with
$n\in\{2,3\}$, which is a common strategy in AId tasks (see for
example the 2019 PAN shared task~\citep{Kestemont:2019ov}). We then
perform feature selection by selecting the $k$ most relevant features
via $\chi^{2}$, with $k=1,000$. We tackle SAV in the style of
\citet{Corbara:2023py}, i.e., we create a single feature vector by
computing the absolute difference among the feature values of the two
documents that make up the document pair, and label the pair as either
\textsc{SameAuthors} or \textsc{DifferentAuthors}.

For the deep-learning experiments, we employ a RoBERTa
model~\citep{Liu:2019ra} from the \textsf{HuggingFace Transformers}
library~\citep{huggingface_transformers} specifically trained with
Latin data.\footnote{Documentation available at:
\url{https://huggingface.co/pstroe/roberta-base-latin-cased3}.} We
fine-tune the model for 5 epochs on the training set, employing the
AdamW optimizer \citep{Loshchilov:2018dd} with initial learning rate
set to $0.0001$ and cross-entropy as the loss. For the SAV task, note
that RoBERTa is able to directly classify a sequence of two texts: it
is sufficient to concatenate the two texts, separated by the
appropriate separator token \textsc{[SEP]}. Note also that RoBERTa
works with a fixed maximum length of 512 tokens; we thus truncate the
textual samples accordingly.

In Table~\ref{tab:eval_results} we report the evaluation results for
each model and for each task. As we can see, both algorithms show very
high performance in all the tasks. Interestingly, while performing
almost on par in the AV task, and better in the SAV task, the RoBERTa
transformer performs slightly worse than the SVM classifier in the
multi-class setting of AA.

\begin{table}[h]
  \centering
  \begin{tabular}{rcccc}
    \toprule
    & \multicolumn{2}{c}{\textbf{SVM}} & \multicolumn{2}{c}{\textbf{RoBERTa}} \\
    \cmidrule{2-5}
    & $Acc$ & $F_{1}$ & $Acc$ & $F_{1}$ \\
    \toprule
    \textbf{SAV} & .836 & .838 & \textbf{.957} & \textbf{.956} \\
    \textbf{AV} & \textbf{.985} & .894 & \textbf{.985} & \textbf{.900} \\
    \textbf{AA} & \textbf{.989} & \textbf{.981} & .978 & .963 \\
    \bottomrule
  \end{tabular}%
  \caption{Evaluation results for the employed classifiers (SVM and
  RoBERTa); the metrics employed are accuracy ($Acc$) and $F_{1}$. For
  SAV and AV, which are binary tasks, $F_{1}$ is defined in the
  standard way, while for AA, which is a single-label multiclass task,
  the reported $F_{1}$ values are obtained by macro-averaging (i.e.,
  they are computed as the arithmetic mean of the class-specific
  $F_{1}$ values). Micro-averaged $F_{1}$ values are not reported
  here, since micro-averaged $F_{1}$ and accuracy are the same measure
  in single-label multiclass classification. The best model result for
  each task is in \textbf{bold}.}
  \label{tab:eval_results}
\end{table}


\section{Tools for explainable authorship identification: A
comparative analysis}
\label{sec:results}

\noindent We here present our results divided by type of explanation,
namely feature ranking (Section~\ref{subsec:results_feats_ranking}),
probing (Section~\ref{subsec:results_probing}), and
factual-counterfactual selection
(Section~\ref{subsec:results_factuals}).


\subsection{Feature ranking for SAV}
\label{subsec:results_feats_ranking}

\noindent In Table~\ref{tab:feature_importance} we show the top five
and bottom five features by coefficient value for the SVM that we have
trained for the SAV task.\footnote{Note that we only show this method
as applied to the SAV case, but the considerations we make here also
apply to AV and AA.}

\begin{table}[]
  \centering
  \begin{tabular}{@{} l r @{}}
    \toprule \textbf{$n$-grams} & \textbf{coef} \\ \midrule
    ``gab'' & 0.193 \\
    ``tto'' & 0.179 \\
    ``mac'' & 0.178 \\
    ``mbi'' & 0.175 \\
    ``aia'' & 0.171 \\
    \multicolumn{2}{c}{\textbf{$\dots$}}\\
    ``auc'' & -1.454 \\
    ``\_ai'' & -1.586 \\
    ``ait'' & -1.725 \\
    ``ae\_'' & -3.976 \\
    ``ae'' & -4.792 \\
    \bottomrule
  \end{tabular}
  \caption{Bottom 5 and top 5 features of the SVM classifier for the
  SAV task by coefficient value (coef). White spaces in the feature
  names are indicated with ``\_''. }
  \label{tab:feature_importance}
\end{table}

In our case, all the feature values are positive, since we employ
TfIdf values, and the intercept is positive as well ($2.29$);
thus, 
features associated with positive weights are important for the
positive class (\textsc{SameAuthor}), and features associated with
negative weights are important for the negative class
(\textsc{DifferentAuthor}).\footnote{The observant reader might get
confused by this notion: since features associated with positive
weights are important for the positive class, and since the feature
values we employ for the SAV task are the result of the absolute
difference among the original feature vectors, does it mean that
higher differences (i.e., higher feature values in the SAV task) are
associated with the two documents sharing the same author? Indeed,
this is counter-intuitive. We can speculate that these features are
not ``significant'' in the common sense, but act as a threshold: in
order for a textual example to be classified as negative, it must have
features values (where the weight is negative) that jointly exceed
these ``insignificant'' feature values (where the weight is
positive). The fact that the positive coefficient values seem
relatively smaller if compared with the negative ones might support
this hypothesis.} Interestingly, we note a disproportion in the
coefficient values among positive and negative weights, where the
positive values appear smaller than the negative ones, slowly
decreasing from the first position toward the value zero. Also, we
note that the two features with the highest negative coefficient
values are ``ae\_'' and ``ae'', meaning that a discrepancy in the
frequency of use of this feature is indeed an indicator that the
authors are different or, put it another way, that the frequency of
use of these features is pretty stable in the production of an author,
and is thus a characteristic trait (either because the author tends to
use it a lot, or only rarely) of an author. This specific case might
be connected with the transitional phase in medieval Latin where
scholars started representing the diphthong ``ae'' with the single
letter ``e'', instead of with the two separate letters ``ae'' as it
was written throughout antiquity. A large difference in the frequency
of use of these features might thus be an indication that the authors
are different, since one of the authors had a preference for ``ae''
while the other instead preferred ``e''.

We can check that the ranked features are indeed useful for the
classification by running an ablation experiment, also known as
\emph{Iterative Removal Of Features} (IROF) \citep{Rieger:2020if}.
Firstly, we assess the performance of SVM on the test set when using
the entire feature set; we then sequentially remove the feature with
the highest absolute coefficient value (by setting the associated
weight to zero) from the feature set, then re-evaluate the model on
the test set (without re-training it).

\begin{figure}[t]
  \begin{center}
    \includegraphics[trim=0 0 0 20,clip, width=.7\linewidth]
    {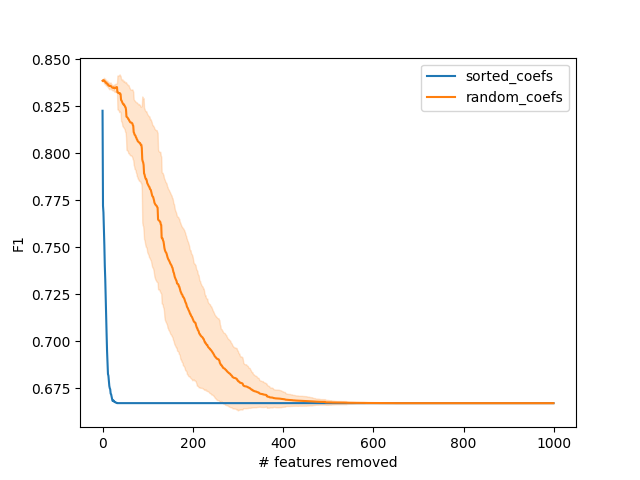}
    \caption{Results of the IROF validation on SVM classifier for SAV:
    we iteratively remove one feature at a time, following the
    descending order of the absolute values of the coefficients
    (\texttt{sorted\_coefs}) or a random feature ranking
    (\texttt{random\_coefs}). In particular, for the latter we show
    the mean $F_{1}$ value obtained at the $n$ feature removed for 10
    random feature rankings, where the coloured shadow is the standard
    deviation.}
    \label{fig:irof}
  \end{center}
\end{figure}

That the model performance drops as we iteratively remove features
should come as no surprise. However, a good ranking of features that
effectively reflects the features importance would cause the
performance to degrade much faster (i.e., in less iterations) than any
other uninformative ranking. This is shown in Figure~\ref{fig:irof},
in which we compare the drop in performance as a function of the
number of features removed, by considering our feature ranking (in
blue) versus (10 trials of) the random ranking (in orange). The fact
that the model becomes a dummy classifier after removing very few
features following our ranking proves that the importance criterion is
indeed informative.

\begin{figure}[t]
  \centering \includegraphics[trim=0 0 0 20,clip, width=.8\linewidth]
  {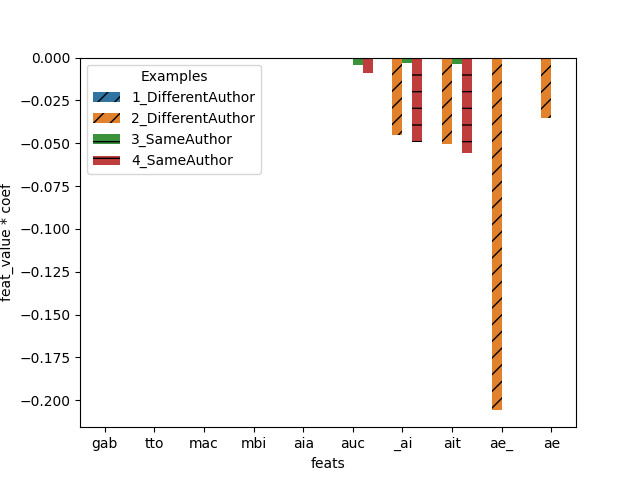}
  \caption{Local explanations for $4$ examples in the test set, given
  the features listed in Table~\ref{tab:feature_importance}; we colour
  the \textsc{SameAuthor} class with the `-' pattern and the
  \textsc{DifferentAuthor} class with the `//' pattern. Note that the
  scores for many features is zero for all four examples. }
  \label{fig:feat_local}
\end{figure}

As already explained in Section~\ref{sec:XAID}, we can also employ the
coefficients to obtain a form of local explanation, by multiplying the
feature value extracted from a document by the correspondent
coefficient. We randomly select $2$ examples for each class
\textsc{SameAuthor} and \textsc{DifferentAuthor}, and show the results
in Figure~\ref{fig:feat_local}. This visualisation highlights the
biggest drawback of this XAI technique when applied to textual
examples: if we limit the investigation to just a few features, we
might risk to convey an incomplete, and thus wrong, picture,
especially to a scholar that is not an expert of machine learning. In
fact, what we observe is that, on the basis of these examples, the
outcome is often contradictory. Firstly, many of the features
displaying positive weights happen to be absent in the selected
examples. Secondly, features displaying negative weights behave
inconsistently across the examples, e.g., showing relatively high
values (examples numbered \textsc{2} and \textsc{4}), or very low
values (examples numbered \textsc{1} and \textsc{3}) regardless of
their class labels (\texttt{SameAuthor} or \texttt{DifferentAuthor}).
Henceforth, 
it seems clear that restricting the study to only a selected number of
features
is not enough to convey the full picture of the model behaviour.

Regarding this XAI approach, we can thus conclude that it contributes
to justifying the decisions of the system in the eyes of the domain
expert to some degree, but it is also rather problematic; as already
noted, AId tasks (as any other application of text classification)
tend to be characterised by a high number of features, each one
providing only a tiny contribution to the final classification
decision. In other words, it is unlikely that there are just a handful
of features that, by themselves, determine a classification
decision. However, as we have shown, limiting the investigation to
only a small portion of the features actually employed by the
classifier incurs in the risk to convey a picture that is just too
narrow and simplistic. Thus, it is of primary importance to offer an
analysis that includes the entirety of the feature set used by the
classifier in the most user-friendly way possible, allowing a scholar
to personalise and navigate the exploration to its full extent.

In particular, a visualisation tool could help display the disputed
document with the occurrences of the different features highlighted,
with the highlighting coming in different shades depending on the
class considered and the significance of the feature. Moreover, the
tool could allow the domain expert to select a particular feature of
interest, and show one or more dataset examples where the feature has
a strong and significant presence.


\subsection{Probing the Transformer for AA}
\label{subsec:results_probing}

\noindent In our experiments, we train a simple Logistic Regression
model as a probe, since the classification head on top of the RoBERTa
transformer has an equivalent complexity, and thus could not gain any
more information from the latent representation.\footnote{In fact,
employing non-linear models as probes could be counterproductive:
their accuracy might be caused by the memorisation of surface
patterns, instead of the information actually captured by the latent
representation~\citep{Hewitt:2019io}.} We take the training set
obtained in Section~\ref{subsec:setup_dataset} and further split it in
a stratified fashion into a training and test set for the probe,
consisting of 90\% and 10\% of the instances respectively. The model
hyperparameters are fine-tuned via 3-fold cross-validation on the
probe training set. The resulting model is then retrained on the full
training set for the probe before evaluation. In our experiments, we
probe the main model trained on the AA task.\footnote{Note that we
only show this method applied to the AA case, but it can be applied to
AV and SAV tasks as well. However, probes for SAV should be handled
carefully, since RoBERTa latent representation involves both texts.
}

As illustrated in Section~\ref{subsec:XAID_probing}, we test the
transformer with five different probings:
\begin{itemize}

\item \textbf{POS chains}: we probe the model for POS chains, where a
  POS chain is a POS $n$-grams with $n\in\{5,10\}$. In particular, the
  probe is issued to predict whether a certain POS chain is present
  ($f(x_i)=1$) or absent ($f(x_i)=0$) in the document; the labelling
  function $f(x_i)$ is thus binary. We extract the POS tags via the
  \textsc{LatinCy} pipeline for the SpaCy
  library.\footnote{Documentation available at:
  \url{https://spacy.io/universe/project/latincy}.} We restrict the
  analysis to the $5$ most discriminative POS chains in the corpus for
  the authors classes, computed via $\chi^2$.

\item \textbf{SQ chains}: the approach is equivalent to the POS chains
  probing. A SQ chain is a SQ $n$-grams with $n\in\{10,15\}$, and we
  extract the syllabic quantities via the prosodic scanner in the
  Classical Language ToolKit (CLTK) library.\footnote{Documentation
  available at: \url{http://cltk.org/}.}
 
\item \textbf{Word lengths}: we probe whether the model takes the
  word-length distribution into account or not. In order to do so, we
  represent each document $x_i$ by means of a histogram
  $(b_i^{(1)},b_i^{(2)}, \ldots, b_i^{(B)})$, in which the bin
  $b_i^{(j)}$ accounts for the relative frequency of words of length
  $j$ (i.e., the fraction of words of exactly $j$ characters) in the
  document. Then, we cluster the documents thus represented in order
  to identify natural groups based on their word-length distribution;
  we use $k$-means as our clustering algorithm and choose the optimal
  number of clusters via the Elbow method within the range
  $[2,10]$. Each cluster is assigned a numerical ID, so that the
  labelling function $f(x_i)$ is categorical in this case, and the
  probe is issued to label each document with the respective $k$-means
  cluster ID. Note that the histogram representation is only used as a
  means for deciding the cluster to which each document belongs; that
  is, the probe is still trained and tested using the internal
  representations $\phi(x_i)$ of the model.\footnote{A technical note:
  we use the implementation of $k$-means provided by the
  \textsf{scikit-learn} library~\citep{scikit-learn}, which relies on
  the Euclidean distance (aka L2) for computing the clusters. This
  turns out to be suboptimal in the case of word lengths, since the
  histograms actually represent ordered distributions, and since the
  L2 does not take into account the order of the dimensions of the
  feature vectors with which it operates. For example, the L2 distance
  between the pair of (normalised) vectors $v_1=(1,0,0,\ldots,0)$ and
  $v_2=(0,1,0,\ldots,0)$ is as large as the L2 distance between the
  same vector $v_1$ and $v_3=(0,0,0,\ldots,1)$, despite the fact that
  $v_1$ and $v_2$ represent documents that tend to use \emph{very
  short} words while $v_3$ instead represents a document that tends to
  use \emph{very long} words. In order to counter this, in this case
  we represent our documents by means of cumulative distributions; in
  our example, this means that the distance between the (cumulative
  distributions) $v'_1=(1,1,1,\ldots,1)$ and $v'_2=(0,1,1,\ldots,1)$
  turns out to be much smaller than the distance between $v'_1$ and
  $v'_3=(0,0,0,\ldots,1)$. A different solution would be to adopt, in
  place of the L2, a distance that is suited for ordinal data, such as
  the Wasserstein distance (aka Earth Mover Distance in computer
  science). However, here we do not explore this possibility since the
  \textsf{scikit-learn} implementation does not allow to customise the
  distance metric.}

\item \textbf{Function words}: we create a probe to check the extent
  to which the model learns from the frequency of use of the function
  words. To this aim, we apply a strategy that is similar to the
  aforementioned case for word lengths. That is, we first represent
  each document $x_i$ as a histogram
  $(b_i^{(w_1)},b_i^{(w_2)}, \ldots, b_i^{(w_B)})$, in which the bin
  $b_i^{(w_j)}$ accounts for the relative frequency of the function
  word $w_j$ in $x_i$. We consider the list of 80 function words for
  Latin used by \citet{Corbara:2023kk}.\footnote{The full list of
  function words is: \textit{a, ab, ac, ad, adhuc, ante, apud, atque,
  aut, autem, circa, contra, cum, de, dum, e, enim, ergo, et, etiam,
  ex, hec, iam, ibi, ideo, idest, igitur, in, inde, inter, ita, licet,
  nam, ne, nec, nisi, non, nunc, nunquam, ob, olim, per, post, postea,
  pro, propter, quando, quasi, que, quia, quidem, quomodo, quoniam,
  quoque, quot, satis, scilicet, sed, semper, seu, si, sic, sicut,
  sine, siue, statim, sub, super, supra, tam, tamen, tunc, ubi, uel,
  uelut, uero, uidelicet, unde, usque, ut}.} As before, we label each
  document with the cluster ID to which it is assigned by a $k$-means
  algorithm based on the histogram-based representations. The function
  $f$ is thus again categorical.

\item \textbf{Genre}: we probe the model for the genre of the
  documents; in particular, we ask the probe to classify the documents
  based on the sub-corpus they belong to, \textsc{MedLatinEpi} or
  \textsc{MedLatinLit}. As such, we try to asses whether the
  transformer encodes the stylistic characteristics of documents of
  epistolary nature ($f(x_i)=1$) versus documents of literary nature
  ($f(x_i)=0$); the labelling function $f(x_i)$ is thus binary.
 
\end{itemize}

\noindent We show the results of the POS probing in the first portion
of Table~\ref{tab:probing_pos_sq}. The probes show high performance
for all the POS chains considered, with the $F_{1}$ performance always
above $0.8$, indicating that the transformer is likely learning from
the syntax of the documents. These results are in line with the
current literature on language model probing
\cite{DBLP:conf/naacl/Liu0BPS19}, and confirm that, even in authorship
analysis, models leverage POS chains in downstream tasks. On the other
hand, the performance of the SQ probing, displayed in the second
portion of Table~\ref{tab:probing_pos_sq}, is quite more limited,
scoring values between $0.6$ and $0.7$. However, these results indeed
show an actual knowledge by the transformer of the concept of syllabic
quantity; this is actually a rather interesting discovery since, to
our knowledge, this is the first work in which this kind of
information is sought in the latent space generated by a transformer.

\begin{table}[t!]
  \centering
  \begin{tabular}{@{} c c c c c c @{}} \toprule & \textbf{chain} &
    $Acc$ & $P$ & $R$ & $F_{1}$ \\ \midrule
    \multirow{5}{*}{\rotatebox{90}{POS}} &
                                           \multicolumn{1}{|l}{adj noun adj noun verb} & .825 & .869 & .825 & .845 \\
                                                & \multicolumn{1}{|l}{adj noun noun adj noun} & .882 & .922 & .882 & .900 \\
                                                & \multicolumn{1}{|l}{adp noun adj noun verb} & .821 & .853 & .821 & .836 \\
                                                & \multicolumn{1}{|l}{noun adj noun adj noun} & .873 & .909 & .873 & .890 \\
                                                &
                                                  \multicolumn{1}{|l}{noun
                                                  adj noun verb verb}
                                                                 &
                                                                   .853
          & .873 & .853 & .863 \\ \midrule & \textbf{chain} & $Acc$ &
                                                                      $P$
                & $R$ & $F_{1}$ \\ \midrule
    \multirow{5}{*}{\rotatebox{90}{SQ}} &
                                          \multicolumn{1}{|l}{$\cup\cup\cup\cup\cup -\cup\cup\cup\cup$} & .670 & .684 & .670 & .674 \\
                                                & \multicolumn{1}{|l}{$\cup\cup\cup\cup\cup\cup\cup - \cup\cup$} & .642 & .647 & .642 & .644 \\
                                                & \multicolumn{1}{|l}{$\cup\cup\cup\cup\cup\cup\cup\cup\cup -$} & .654 & .664 & .654 & .657 \\
                                                & \multicolumn{1}{|l}{$\cup\cup\cup\cup\cup\cup\cup\cup\cup\cup$} & .601 & .614 & .601 & .601 \\
                                                & \multicolumn{1}{|l}{$\cup\cup\cup\cup\cup\cup\cup\cup\cup\cup\cup$} & .626 & .670 & .626 & .639 \\
    \bottomrule
  \end{tabular}
  \caption{POS and SQ probes results. Probes try to predict the
  presence of the given \textbf{POS $n$-gram} or \textbf{SQ $n$-gram}
  in the latent representation of the model. Note that for SQ we here
  employ the standard notation where $\cup$ stands for a short
  syllable and $-$ stands for a long syllable.}
  \label{tab:probing_pos_sq}
\end{table}

Regarding word lengths and function words, we show the results of the
two multi-class classifications in the first and second portion of
Table~\ref{tab:probing_wordlength_fw} respectively; interestingly, the
optimal number of clusters is $6$ for both experiments. The probe show
poor or mediocre results, getting higher scores in inferring the
function-words distribution of the documents. This highlights the
importance of elements such as function words in the characterisation
of literary authors~\citep{Kestemont:2014uc}. We can hypothesise that
the fact that the transformer does not seem to encode the information
regarding word-lengths distribution could be due to the authors having
similar backgrounds, and thus similar habits regarding their
vocabulary usage.

\begin{table}[t!]
  \centering
  \begin{tabular}{@{} l c c c c c @{}} \toprule & \textbf{\#clusters}
    & $Acc$ & $P$ & $R$ & $F_{1}$ \\ \midrule \textbf{Word lengths} &
    6 & .487 & .487 & .487 & .486 \\ \midrule
    \textbf{Function words} & 6 & .617 & .628 & .617 & .617 \\
    \bottomrule
  \end{tabular}
  \caption{Word-lengths and function-words probes results. Probes try
  to predict the \textbf{word-lengths cluster} or \textbf{function
  -words cluster} in the latent representation of the model. }
  \label{tab:probing_wordlength_fw}
\end{table}

Regarding probing for the genre of the documents, the results can be
seen in Table~\ref{tab:genre}. The probe is clearly able to determine
the sub-corpus the document under consideration comes from, thus
indicating that the transformer indeed encodes the genre of the
document into the latent space. This result could help warn the human
expert against the risk that the neural model under investigation
might be exploiting domain information, which should be avoided in AId
studies \citep{Bischoff:2020mp, Halvani:2019sn}: the classifier should
be focusing on style-related information, and not labelling a document
as written by author $A$ simply on the ground that $A$ often writes in
the same genre or topic as the document in question.

\begin{table}[t!]
  \centering
  \begin{tabular}{@{} l c c c c c @{}} \toprule & $Acc$ & $P$ & $R$ &
    $F_{1}$ \\ \midrule
    \textbf{Genre} & .979 & .979 & .979 & .979 \\
    \bottomrule
  \end{tabular}
  \caption{Genre probes results. Probes try to predict the
  \textbf{sub-corpus} of the documents among \textsc{MedLatinEpi} and
  \textsc{MedLatinLit} in the latent representation of the model. }
  \label{tab:genre}
\end{table}

Summing up, 
this analysis could indeed reveal to a scholar some of the inner
workings of a high-level model, by showing which features it leverages
and which it avoids, thus reassuring the scholar of the outcome of the
classification. In particular, the probing task would
be 
well suited for an active interaction with the 
scholar who, prompted by their deep knowledge on the literary matter,
could propose promising features to analyse, in the form of
`human-in-the-loop'' process. However, an important limitation
scholars should be aware of, is that the probes can only be
constructed around automatically decidable features, unless one wants
to incur the cost of manually labelling the documents according to
more complicated features.


\subsection{Factuals and counterfactuals for AV}
\label{subsec:results_factuals}

\noindent In our experiments, we retrieve one factual and one
counterfactual for both SVM and RoBERTa trained for the AV
task\footnote{Note that we only show this method as applied to the AV
case; however, it can be applied to the SAV and AA tasks as well.  }
by using the Euclidean distance. Specifically, we obtain the TfIdf
vectors (in the SVM case) or the encodings of the final hidden state
(in the RoBERTa case) for a random test instance $x$ and the entire
training set; we then compute the distances among $x$ and all the
training instances, and select the training instance closest to $x$,
which has the same (different) label as the predicted label of $x$. Of
course, both the number of (counter)factuals outputted and the
similarity measure to employ are parameters that can be modified.

The selected test example, which is an epistle from the author Pier
della Vigna, is the following:

\begin{quote}
  Fridericus uniuersis mundi \textit{\textcolor{red}{pri}ncipibus} de
  \textcolor{blue}{ si}nistris rumoribus \textbf{Terrae Sanctae} Etsi
  \textcolor{blue}{tam} iusta quam uehemens \textcolor{blue}{ ca}usa
  doloris et motus fuerit in nobis cum ad presentiam nostram
  \textbf{frater S.\textcolor{violet}{ a }ue\textcolor{blue}{ner}abili
  p\textcolor{red}{atr}e p\textcolor{red}{atr}iarcha} Antiocheno
  dilecto \textit{amico nostro} presentium baiulus litterarum
  acc\textcolor{violet}{ess}it ipsum \textcolor{blue}{tam}en infeste
  uidere nequiuimus qui mittentem affectione quadam diligimus
  \textcolor{blue}{ si}\textcolor{violet}{ng}ulari. Uerum etiam tunc
  temporis cordis nostri \textcolor{blue}{ner}uum
  perti\textcolor{violet}{ng}erat rumor infestus et subitae nuntius
  tempestatis qui Coheminorum pestem ab originalibus sedibus Tartarea
  clade depulsam uelut molem i\textcolor{violet}{ng}entem per abrupta
  montium et decliuium fulminis ictibus
  d\textcolor{red}{eu}olu\textcolor{blue}{tam} in
  \textbf{Sanc\textcolor{blue}{tam} Ciuitatem} irruisse crudeliter
  nuntiaui\textcolor{red}{t. }Qua\textcolor{red}{e f}orte desolationis
  suae tempore habitatore continui solita defensari \textcolor{blue}{
  ca}teruatim undique concurrentibus \textit{populis} colebatur
  dederatque cursui famosi \textcolor{blue}{tam}en loci
  lo\textcolor{violet}{ng}is retro temporibus\textcolor{violet}{ a }
  \textbf{Christicolis} maxime desiderata securitas et
  \textcolor{blue}{ si}nistris \textbf{\textcolor{red}{ au}spiciis}
  diebus \textcolor{red}{ill}is obtenta quarumdam occasione
  \textit{tr\textcolor{red}{eu}garum} quas \textit{soldanus} Damasci
  et Nathasar \textit{soldanus} Cr\textcolor{blue}{aci} qui
  \textcolor{red}{pri}us \textit{hostes} et \textit{aduersarii}
  fuerant \textit{concordiam} inuicem f\textcolor{blue}{aci}entes
  ipsam cum \textbf{Christianis} ea condicion\textcolor{red}{e
  f}ecerunt quod tota regni Hie\textcolor{blue}{ros}olimitani terra
  quam \textbf{Christiani} possederant trans Iordanem
  retentis\textcolor{blue}{ si}bi u\textcolor{red}{ill}is et montanis
  aliquibus restituta \textbf{Christiani} \textit{soldanis} eisdem in
  expugnatione \textit{soldani} Babiloniae deberent assistere toto
  posse. Qua \textit{confederatione} \textcolor{blue}{tam}quam in sui
  perniciem inita \textit{soldanus} accinctus
  predic\textcolor{blue}{tam} gentem \textit{Barbaricam} Coheminorum
  per deserta uagantem et uelut feram in saltibus ante uenabulum
  fugientem ad suae defensionis \textit{\textcolor{red}{ au}xilium}
  conuocaui\textcolor{red}{t. }qui\textcolor{blue}{ si}bi reputantes
  oblatum presidium potrius quam petitum ad designata loca subito non
  minus t\textcolor{blue}{aci}ti quam celeres
  perue\textcolor{blue}{ner}unt ut inuisos \textit{hostes} aduenisse
  maturis nostrorum uigilantia nouerit quam
  uentu\textcolor{blue}{ros}. \textcolor{blue}{
  si}cqu\textcolor{red}{e f}actum est ut \textbf{Christianorum}
  excercitu cum soldanis predictis in guerram soldani Babiloniae apud
  Gazaram commorante \textbf{p\textcolor{red}{atr}iarcha}
  Hie\textcolor{blue}{ros}olimitanus de partibus Cismarinis ad partes
  \textcolor{red}{ill}as athleta nouus
  acc\textcolor{violet}{ess}i\textcolor{red}{t. }[\dots]
\end{quote}

\noindent It is the narration of an episode of the crusade in the Holy
Land, with the characteristics of historical chronicles; it is one of
the many letters that the author wrote while chancellor of the Emperor
Frederick II.

Both SVM and RoBERTa model hold the same factual example, which is
again an epistle, but curiously from the author Giovanni Boccaccio:

\begin{quote}
  Celeberrimi nominis \textit{militi} Iacobo
  Pizi\textcolor{violet}{ng}e serenissimi \textit{principis} Federici
  Trinacrie regis logothete. Gene\textcolor{blue}{ros}e \textit{miles}
  incertus mei Neapoli aliquamdiu fueram uere preterito. hinc enim
  plurimo desiderio trahebar redeundi in \textit{patriam} quam
  autumpno nuper elapso indignans liqueram nec minus reuisendi
  libellos quos immeritos omiseram \textcolor{blue}{ si}c et
  \textit{amicos} aliosque \textcolor{blue}{
  ca}\textcolor{blue}{ros}. inde uero urgebar ut consisterem atque
  detinebar nunc\textcolor{violet}{ a }ue\textcolor{blue}{ner}abili
  uiolentia nunc suasionibus nunc precibus incliti uiri Hugonis de
  \textit{comitibus} \textbf{Sancti} Seuerini cuius credo splendidam
  famam noueris. Curabat enim uir eximius etiam me inuito totis
  uiribus ut me interueniente subsidio serenissime \textit{domine}
  Iohanne Ierusalem et \textcolor{blue}{ si}cilie \textit{regine} apud
  Parthenopeos pl\textcolor{blue}{aci}do locaret in otio. qua
  perplexitate a\textcolor{violet}{ng}ebar nimium nulla adhuc in parte
  satis firmato consilio. Et dum\textcolor{blue}{ si}c uariis agitarer
  curis quo pacto non memini factum \textcolor{blue}{tam}en est ut ad
  aures deueniret meas ue\textcolor{blue}{ner}abile nomen
  \textbf{religiosi} hominis Ubertini de \textbf{ordine} Minorum
  \textbf{sacre theologie} prof\textcolor{violet}{ess}oris et conciuis
  tui cuius auditis meritis eumque ea tempestate Neapoli moram trahere
  pro quibusdam arduis tui suique \textit{regis} in desiderium uenit
  \textcolor{blue}{tam} conspicuum uidere uirum. \textcolor{violet}{ a
  }pueritia quippe mea etiam ultra tenelle etatis uires talium
  auidissimus fui. Nec mora. exhibiturus \textit{reuerentiam}
  debi\textcolor{blue}{tam} ad eum acc\textcolor{violet}{ess}i atque
  adaperto \textcolor{blue}{ ca}pite primo paxillum miratus hominem
  quam \textbf{deuotissime} et humillime potui salutaui eum. Ipse
  autem graui quadam maturitate obuius factus me leta
  f\textcolor{blue}{aci}e miti eloquio et morum laudabili comitate
  suscepit.
\end{quote}

\noindent In this case, the selected factual is a letter to
(ironically!) a notary of the Kingdom of Sicily; the epistle presents
a first-person narration, describing some personal occurrences of the
author during a staying in Naples. The themes apparently could not be
more different from the test example, but at a closer inspection the
two texts share many references to religious orders and political
relations. We demonstrate this by formatting in \textbf{bold} some of
the former and in \textit{italic} some of the latter.

Regarding the counterfactual, SVM and RoBERTa again hold the same
example, which is yet another epistle, this time form the author Dante
Alighieri (of course, since in this experiment Dante is the positive
class, while all the other authors are the negative class):

\begin{quote}
  Absit\textcolor{violet}{ a }uiro predicante iustitiam ut
  perp\textcolor{violet}{ess}us iniurias iniuriam inferentibus uelut
  benemerentibus pecuniam suam solua\textcolor{red}{t. }Non est hec
  uia red\textcolor{red}{eu}ndi ad \textit{p\textcolor{red}{atr}iam}
  pater mi. sed si alia per uos ante\textcolor{red}{ au}t deinde per
  alios inuenitur qu\textcolor{red}{e f}ame Dantisque \textit{honori}
  non deroget \textcolor{red}{ill}am non lentis passibus
  acceptabo. quod si per nullam talem Florentia introitur nunquam
  Florentiam introibo. Quidni. nonne solis astrorumque specula ubique
  conspiciam. nonne dulcissimas ueritates potero speculari ubique sub
  celo ni \textcolor{red}{pri}us i\textcolor{violet}{ng}lorium ymo
  ignominiosum \textit{populo} Florentino \textit{ciuitati} me
  reddam. Quippe nec panis deficiet.
\end{quote}

\noindent Unlike the factual, the counterfactual clearly has a very
different domain than the test sample. It is a personal account of the
tribulations of returning (unwanted) to one's homeland. Still, there
are again some references to political concepts (again shown in
\textit{italic}).

All in all, it seems that both models are able to spot similarities
and differences in the documents, especially the ones linked with the
themes and references of the narration. However, spotting these
similarities and differences appears to be mainly in the hands of the
human user, which could be a long and difficult task. Coupling this
XAI technique with other methods that allow to simultaneously
enlightening the textual regions, or features, that most determine the
similarity among the documents could be helpful in this sense. We give
a very elementary exemplification of this by colouring in the texts
the $10$ $n$-grams that have the minimum differences among the
features values in the test example and in the factual (in
\textcolor{blue}{blue}), and among the features values in the test
example and the in counterfactual (in \textcolor{red}{red}) (the
$n$-grams that are shared among all three texts are coloured in
\textcolor{violet}{violet}).

Other techniques that exist in the related literature generate
ad-hoc synthetic examples as (counter)factuals (see for example
\citet{Lampridis:2022ge}) . 
While it might be possible in principle to generate synthetic textual
instances for our case too,
it is not clear how
these examples could be useful for the human expert, who would
realistically be interested in real-world textual documents, not
machine-generated texts.
 

\section{Conclusion and future works}
\label{sec:conclusion}

\noindent In this article, we underline the importance of
explainability for authorship studies, with a specific focus on the
case of cultural heritage.
Despite its importance, there are no existing XAI techniques
particularly devised for authorship studies in the field of cultural
heritage, nor even for generalised applications of authorship
analysis. We thus experiment with three existing XAI methodologies
proposed in other contexts (namely, feature ranking, probing, and
factual and counterfactual selection), and we test them on the three
main Authorship Identification tasks (Authorship Attribution,
Authorship Verification, and Same-Authorship Verification), employing
a medieval Latin dataset as case-study. We make the code here
developed available to other researchers that might want to apply
these techniques to other authorship problems.

We show that each of the XAI methods partially contributes to
understand the reasons behind the predictions of the model, and they
jointly provide some sort of explanations of different aspects of the
model.  In particular, while features ranking and probing shed light
on the linguistic events that are leveraged by the
model, 
(counter)factuals put these important linguistic events in context,
showing real examples of the writing production under study.

However, we argue that the explanations that can be obtained with
current, general-purpose techniques are still largely insufficient,
since they either convey a rather limited perspective of the inner
working of the classifier (feature ranking) or heavily rely on the
user input and intuition (probing, factuals and
counterfactuals). Employing a combination of these methods, instead
than using them in isolation, would mitigate, but not resolve, this
obstacle.

In future work, the exploration for suitable methods to provide
meaningful explanations for supporting the research of scholars should
continue. We believe that targeting concise and informative textual
explanations is a promising way worth exploring, since this is the
format most familiar to cultural heritage scholars (see for example
\citet{Barratt:2017te, Le:2020uj}).


\subsubsection*{Acknowledgments}

\noindent The work by Alejandro Moreo and Fabrizio Sebastiani has been
supported by the \textsf{SoBigData++} project (Grant 871042) and by
the \textsf{AI4Media} project (Grant 951911), both funded by the
European Commission (Grant 871042) under the H2020 Programme, and by
the \textsf{SoBigData.it} and \textsf{FAIR} projects, both funded by
the Italian Ministry of University and Research under the
NextGenerationEU program. The authors' opinions do not necessarily
reflect those of the funding agencies.



\end{document}